\documentclass[letterpaper, 10 pt, conference]{ieeeconf}  

\IEEEoverridecommandlockouts                             

\usepackage{graphicx} 
\usepackage{mathtools}
\usepackage{amsmath}
\usepackage{bm}
\usepackage{xcolor}
\usepackage{lipsum}
\usepackage{amssymb}  

\usepackage[style=ieee]{biblatex}

\DeclareSourcemap{
  \maps{
    \map{
      \pertype{article}
      \step[fieldset=url, null]
      \step[fieldset=doi, null]
      \step[fieldset=issn, null]
      \step[fieldset=isbn, null]
      \step[fieldset=note, null]
      \step[fieldset=editor, null]
      \step[fieldset=urldate, null]
      \step[fieldset=file, null]
    }
  }
}
\DeclareSourcemap{
  \maps{
    \map{
      \pertype{inproceedings}
      \step[fieldset=url, null]
      \step[fieldset=doi, null]
      \step[fieldset=issn, null]
      \step[fieldset=isbn, null]
      \step[fieldset=note, null]
      \step[fieldset=editor, null]
      \step[fieldset=urldate, null]
      \step[fieldset=file, null]
    }
  }
}
\DeclareSourcemap{
  \maps{
    \map{
      \pertype{incollection}
      \step[fieldset=url, null]
      \step[fieldset=doi, null]
      \step[fieldset=issn, null]
      \step[fieldset=isbn, null]
      \step[fieldset=note, null]
      \step[fieldset=editor, null]
      \step[fieldset=urldate, null]
      \step[fieldset=file, null]
    }
  }
}

\title{\LARGE \bf
Enabling steep slope walking on Husky using reduced order modeling and quadratic programming
}

\author{Kaushik Venkatesh Krishnamurthy$^1$, Eric Sihite$^2$, Chenghao Wang$^1$, Shreyansh Pitroda$^1$, \\ Adarsh Salagame$^1$, Alireza Ramezani$^{1*}$, and Morteza Gharib$^2$ 
\thanks{$^{1}$The authors are with the Silicon Synapse Labs, Department of Electrical and Computer Engineering,
        Northeastern University, Boston, MA, USA  {\tt\small venkateshkrishnamu.k, wang.chengh, pitroda.s, a.salagame@northeastern.edu}}%
\thanks{$^{2}$The authors are with the Department of Aerospace Engineering, California Institute of Technology, Pasadena, CA, USA {\tt\small esihite, mgharib@caltech.edu}%
}
\thanks{$^{*}$Corresponding author {\tt\small a.ramezani@northeastern.edu}}}

\begin{document}

\maketitle
\thispagestyle{empty}
\pagestyle{empty}

\begin{abstract}
Wing-assisted inclined running (WAIR) observed in some young birds, is an attractive maneuver that can be extended to legged aerial systems. This study proposes a control method using a modified Variable Length Inverted Pendulum (VLIP) by assuming a fixed zero moment point and thruster forces collocated at the center of mass of the pendulum. A QP MPC is used to find the optimal ground reaction forces and thruster forces to track a reference position and velocity trajectory. Simulation results of this VLIP model on a slope of 40 degrees is maintained and shows thruster forces that can be obtained through posture manipulation. The simulation also provides insight to how the combined efforts of the thrusters and the tractive forces from the legs make WAIR possible in thruster-assisted legged systems.
\end{abstract}

\section{Introduction}
\label{sec:intro}

Wing Assisted Inclined Running (WAIR) is a maneuver that was first introduced and studied in \cite{dial_wing-assisted_2003,tobalske_aerodynamics_2007,peterson_experimental_2011}, and was performed by young birds while climbing up steep slopes. The authors' investigation found that the wings along with the legs were essential in the WAIR maneuver to provide thrust and traction respectively, enabling vertical running up trees and consequently the ability to forage and escape predation at a very young age.

\begin{figure}[t]
    \centering
    \includegraphics[width=\linewidth]{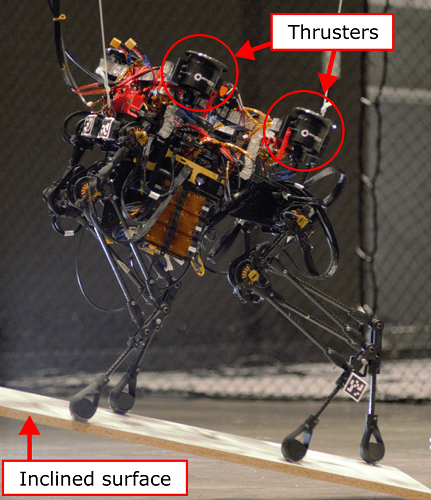}
    \caption{Image of Husky Carbon \cite{ramezani_generative_2021}, a legged-aerial robot, positioned on an inclined surface. Husky features four electric ducted fans mounted on its torso and twelve custom-designed actuators with high energy density.}
    \label{fig:cover-image}
\end{figure}

In this research, we aim to mimic this capability on the multi-modal legged aerial platform Husky (shown in Fig.~\ref{fig:cover-image}) using a reduced-order model and an optimal control problem. The Northeastern University Husky Carbon \cite{ramezani_generative_2021, sihite_optimization-free_2021, salagame_quadrupedal_2023,sihite_dynamic_2023,liang_rough-terrain_2021,salagame_quadrupedal_2023-1,salagame_letter_2022,krishnamurthy_thruster-assisted_2024,krishnamurthy_thruster-assisted_2024,krishnamurthy_narrow-path_2024} has been conceived and developed at Northeastern University in Boston. The robot weighs 8 kilograms and stands ~0.6 meters tall and about ~0.3 meters at its widest. Husky is purpose-built to have both flight and walking capabilities, where both modes have conflicting ideal requirements.  These problems were circumvented by building a lightweight structure and designing actuators with harmonic drives with 3D-printed housing.  Each leg boasts 3 DOFs with actuators which are powered by high-power ELMO amplifiers, enabling torque control through current regulation in the DC actuator windings. In total the legged system has 12 actuated DOFs. To fly, Husky is fitted with powerful propellers on its back. With its hardware and the ability to provide thrust forces using posture manipulation, WAIR becomes an attractive maneuver to pursue on Husky, with strong implications in legged-aerial systems through thruster-assisted locomotion.

The goal of the Husky Carbon Platform has been to explore multi-model legged locomotion through posture manipulation and appendage repurposing \cite{sihite_dynamic_2023, sihite_multi-modal_2023, mandralis_minimum_2023, sihite_demonstrating_2023}. Additionally, the goal is also to push the boundaries beyond standard legged locomotion and, by doing so, being able to exploit the most out of each modality's advantages in legged-aerial robots. Most legged robots have the ability to only manipulate the ground contact forces using posture manipulation making dynamic maneuvers on steep slopes hard. The strategies and abilities of robots walking on slopes have been reported in \cite{ma_quadrupedal_2020, xin_optimization-based_2024, peterson_wing-assisted_2011, roennau_reactive_2014, gehring_dynamic_2015}, showing varied methods to tackle steep slopes. The TITAN \cite{komatsu_how_2015, hirose_titan_1997} robots show legged locomotion on steep slopes using a gait with legs that sprawl out from the body and powerful actuators to actuate the legs which carry the weight of the robot. Komatsu et al.~\cite{komatsu_how_2015} present a method to optimize slope walking with the TITAN robots and find walking postures and actuator constraints to improve energy efficiency. Ma et al. ~\cite{ma_quadrupedal_2020} show trotting on slopes up to 25 degrees with a quadrupedal robot which uses a collocation-based optimal controller and decomposing the quadrupedal dynamics into coupled dynamics of bipedal robots. Focchi et al.~\cite{focchi_high-slope_2017} and earlier studies by Hutter et al.~\cite{hutter_hybrid_2012} exhibit impressive walking on steeply sloped surfaces while maintaining friction cone constraints using a robust and effective QP-based force distribution controller for challenging terrain. Katz et al. \cite{katz_mini_2019} show the efficacy of QP based MPC solvers to perform various dynamic maneuvers. 

Some robots use inherent morphology \cite{griffin_walking_2017} to generate angular momentum to balance on narrow platforms while, on the other hand, some robots have additional mechanisms such as reaction wheel actuators \cite{lee_enhanced_2023}, to generate the required momentum to balance from external disturbances and narrow platforms. Unfortunately, these solutions are incompatible with the conflicting requirements of a legged aerial system

\begin{figure*}[t]
    \includegraphics[width=\linewidth]{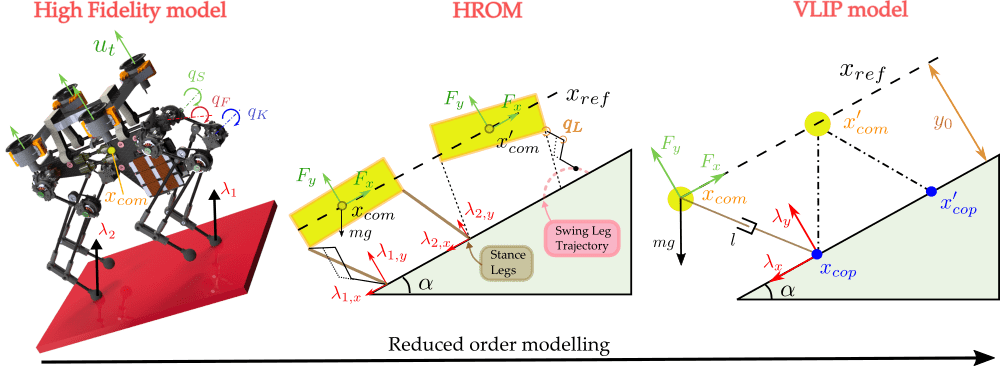}
    \caption{Depicts the parameters of the full, Husky reduced-order model (HROM), and the planar variable length inverted pendulum (VLIP) model, which are employed to govern the equations of motion as detailed in Section~\ref{sec:modelling}. The thruster forces and moments are applied at the body's center of mass to assist in walking and climbing up the slope.}
    \label{fig:wair}
\end{figure*}

However, the aspects of multi-modal legged aerial robots that use thrust vectoring through posture manipulation for locomotion are relatively unexplored. The challenges involve maintaining the kinematic constraints during posture manipulation for the purposes of thrust vectoring, and friction cone constraints and. Earlier studies show locomotion up slopes but they only consider the walking gait where 3 feet are always in contact with the ground. We aim to realize an agile and dynamic two-point contact gait with Husky while using thruster forces and a reduced order model that can be tractable and scaled to real-time control schemes.

Using a full-order model or a rigid body model with massless legs with no-slip constraint for foot ends can help create models that can closely estimate the ground reaction force on each leg given the state, joint trajectories, and the thruster input. This allows us to formulate a controller to find thruster forces within the friction cone and the constraint admissible set. However, these models are nonlinear and can get very complicated with multiple constraint equations. Furthermore, certain configurations can sometimes lead to cases where the Jacobians for the constraint matrices are not full rank thus making it impossible to find individual ground reaction forces or very slow with least square estimations.

To this extent, we introduce a modified Variable Length Inverted Pendulum (VLIP) as a walking model for WAIR. VLIP models have been used extensively in history for legged systems \cite{kim_bipedal_2021}. This model includes an external force on the point mass of the pendulum to represent the thruster forces that the actual robot can produce. By considering the VLIP model as constrained to a ZMP point on the slope, which is defined at a fully controllable COP point, the ground reaction forces that obey friction cone conditions are found. The thruster forces that obey the full equations of motion of the pendulum point mass then help make the trajectories of the VLIP possible. 

To track a reference trajectory optimally and find the constraint-admissible ground forces, we formulate an MPC with a quadratic cost and use Quadratic Programming (QP) tools to exploit the inherent simplicity of minimizing quadratic cost functions. This study uses \textit{qpSWIFT} \cite{pandala_qpswift_2019} which is a lightweight Primal-Dual interior point solver and employs sparsity properties that enable solving optimization problems in real-time. This solver has been used with extensive success for optimal control and trajectory tracking for legged robots which are tractable in real-time, by Ding et al. in the following studies \cite{ding_representation-free_2021, ding_real-time_2019}.  

The contributions for this paper are then as follows:
\begin{enumerate}
    \item First, WAIR, a maneuver involving thruster-assisted dynamic legged locomotion on steep slopes, is described using a reduced-order model based on VLIP. While VLIP itself is not a new concept in legged locomotion, integrating external thruster actions to simulate WAIR represents a new approach.
    \item A QP-based MPC to find optimal, constraint-admissible reaction forces and thruster forces is proposed and simulated.
\end{enumerate}

The structure of the paper is as follows. In Section 1, the paper explains the dynamic modelling of HROM and in Section 2 follow it up with the modified VLIP model and outline its equations of motion. In Section 3, the paper outlines the QP MPC and in Section 4 validate the stability of our controller through MATLAB simulations of the VLIP on various slopes. This work is concluded with final remarks and future directions.

\section{Dynamics Modelling}
\label{sec:modelling}

The dynamics model for the simulation follows the Husky reduced-order model (HROM). This model is defined with a single inertial body and massless legs where a thruster wrench can be applied at the center of mass and ground forces are applied at the foot that's in contact with the ground. 

The generalized coordinates of the robot body can then be defined as follows:
\begin{equation}
\bm{q} = \left[\bm{p}_B^\top ,\bm{\Phi}_B^\top \right]^\top 
\label{eq: body states}
\end{equation}
where $\bm p_B$ and $\bm \Phi_B$ are the body inertial position and Euler angles, respectively. The massless leg states of the robot can be defined as,
\begin{equation}
\begin{aligned}
\bm{q}_L &= \left[\dots, \phi_i,\gamma_i,l_i, \dots\right]^\top, \quad
i \in \mathcal{F},
\label{eq: leg states}
\end{aligned}
\end{equation}
where $\phi_i$, $\gamma_i$, and $l_i$ are the leg $i$'s hip frontal and sagittal joint angles, and the length, respectively. Furthermore, $\mathcal{F} =\left\{FR, HR, FL, HL\right\}$ represents the set of four legs (combinations of front/hind and right/left). The inertial position of the foot can then be determined using the following forward kinematics equations:
\begin{equation}
\begin{aligned}
    \bm{p}_{f i} &= \bm{p}_{B} + R_{B} \bm{l}_{h i}^{B} + R_{B} \bm{l}_{f i}^{B} \\
    \bm{l}_{f i}^{B} &= R_{y}\left(\phi_{i}\right) R_{x}\left(\gamma_{i}\right)
    \begin{bmatrix} 
    0 & 0 & -l_{i}
\end{bmatrix}^\top
\label{eq:foot_pos}
\end{aligned}
\end{equation}
where $R_B(\bm \Phi_B)$ is the rotation matrix from the body to the inertial frame and the superscript $B$ denotes a vector defined in 
the body frame.
The inertial positions of the thrusters ($\bm{p}_{ti}$) are defined as follows:
\begin{equation}
    \bm{p}_{ti} = \bm{p}_B + R_B\bm{l}_{t,i}^B
\end{equation}
As the legs are considered massless, the kinetic and potential energies of the HROM can be calculated using the equations provided below:
\begin{equation}
\begin{aligned}
    \mathcal{K} &= \left( \frac{1}{2} \dot{\bm{p}}_B m \dot{\bm{p}}_B^\top +\bm{\omega}_B^B I_B \bm{\omega}^{B\top}_B \right) \\
     \mathcal{V} &= -m \bm{p}_B^\top \bm{g} \\
    \mathcal{L} &= \mathcal{K}-\mathcal{V},
\label{eq:LKV}
\end{aligned}
\end{equation}
where $m$ and $I_B$ are the body's mass and inertia matrix, $\bm{\omega}_B^B$ represents the body angular velocity in the body frame, and $\bm{g}$ denotes the gravitational acceleration vector. The angular velocity of the body can be found as a function of the rate of change of Euler angles using the Euler rate matrix $E(\bm{\Phi}_B)$,
\begin{equation}
\bm{\omega}_B^B = E(\bm{\Phi_B}) \dot{\bm{\Phi}}_B
\end{equation}
 The dynamic equation of motion can be derived using the Euler-Lagrangian method as follows:
\begin{equation}
    \textstyle \frac{d}{dt}\left(\frac{\partial{\mathcal{L}}}{\partial{\bm{v}}}\right)- \frac{\partial \mathcal{L}}{\partial{\bm{q}}} = \bm{\Gamma},
\label{eq:euler-lagrangian}
\end{equation}
where $\bm v = \dot{\bm q}$ contains the generalized velocities and $\bm{\Gamma}$ is the sum of all generalized torques and forces acting on the inertial body. The dynamic system accelerations can then be solved following the standard form:
\begin{equation}
    \bm M\dot{\bm v} + \bm h = \bm u_e + \sum_{i \in \mathcal{F}}\bm J_{i,l}(\bm q_d
    )^{\top} \bm \lambda_i,
    \label{eq: eom}
\end{equation}
where $\bm M $ is the mass/inertia matrix, $\bm h$ contains the Coriolis and gravitational vectors, and $\bm{u}_e~\in~\mathbb{R}^6$ is the external thrust wrench acting on the COM of the rigid body of the HROM by the thrusters. $\bm q_d$ contains $\bm q$ and leg state positions.
The equations of motion can then be written in the following form:
 \begin{equation}
\begin{gathered}
    \dot{\bm{x}} = \bm{f}(\bm{x},\bm{u},\bm{\lambda}), \\
    \bm{x} = \left[\bm q_d^\top ,\dot{\bm{q}}_d^\top \right]^\top\\
    \bm{u} = \left[\bm{u}_e^\top, \bm{u}_L^\top\right]^\top\\
    \end{gathered}
\label{eq:eom}
\end{equation}
%
The no-slippage condition is ensured since in the subsequent step, outlined in Section~\ref{sec:mpc}, we obtain the control actions that maintain the states within the constrained-admissible set.
\section{WAIR Reduced-Order Model}
\label{sec:LIP_model}
To reduce computation time and present a simpler model for the MPC to solve, we derive a simplified planar model based on a linear inverted pendulum (LIP) model (see Fig.~\ref{fig:wair}). The model plane is defined to be about the vertical axis and the direction of the incline. The planar center of pressure (COP) equation of motion can be defined as follows:
\begin{equation}
\begin{aligned}
    m\ddot{x} &= -\lambda_x - mg \sin\alpha + F_x\\
    m\ddot{y} &= \lambda_y  - mg \cos\alpha + F_y 
\end{aligned}
\label{eq:planar_eom}
\end{equation}
where $m$ is mass, $F_x $ and $F_y$ are the Thruster forces, $\alpha$ is the slope angle, and $\lambda_x$ and $\lambda_y$ are the ground reaction forces. Here, the $x$ position is parallel to the incline direction while $y$ is perpendicular to the incline. 

We assume that the COP position is known and can be regulated to any position inside the support polygon by distributing the forces between the two legs in contact. Solving zero moment point (ZMP) about the center of mass (COM) results in the following equation:
\begin{align}
    \left(\ddot{x}_{com} + \frac{\lambda_x}{m} \right) y_0 = \left( \ddot{y} - \frac{\lambda_y}{m}\right) (x_{cop} - x_{com})
\label{eq:zmp_com}
\end{align}
$x_{com}$ is the position of the center of mass, $x_{cop}$ is the position of the center of pressure, and $y_0$ is the perpendicular height of the center of mass from the slope, which in LIP model is assumed to be constant ($\ddot{y} = 0$). Therefore, \eqref{eq:zmp_com} can be redefined as 
\begin{equation}
\begin{aligned}
    \ddot{x}_{com} &= - \frac{(x_{cop} - x_{com})\lambda_y}{my_0} -\frac{\lambda_x}{m}
\end{aligned}
\label{eq:lip_eom}
\end{equation}

Define $\bm x = [x_{com}, \dot x_{com}]^\top$ and $\bm u = [\lambda_x, \lambda_y]^\top$.
Then, following \eqref{eq:lip_eom}, the dynamical equation of motion is reduced to the following linear time-invariant model:
%
\begin{equation}
\begin{aligned}
    \dot{\bm x} = \underbrace{\begin{bmatrix}
        0 & 1 \\
        -\frac{\lambda_{y0}}{m y_0} & 0 
    \end{bmatrix}}_{\bm A} \bm x  + \underbrace{\begin{bmatrix}0 & 0 \\ - 1 & \frac{-(x_{cop,0} - x_{com,0})}{y0} \end{bmatrix}}_{\bm B} \bm u
\end{aligned}
\label{eq:ss_lip}
\end{equation}
The inputs to this model are ground reaction forces ($\lambda_x$ and $\lambda_y$), which will be solved by the QP solver for the MPC. Upon observation of \eqref{eq:zmp_com}, it can be seen that the VLIP dynamics and the state model, do not depend on the incline angle of the slope, but rather depends on the relative position of the COM with respect to the COP. 

The VLIP dynamics uses ground reaction forces to propel the body using traction forces up the slope and the corresponding thruster forces (which is a function of the body acceleration and incline angle) can be found by solving for $\ddot x$ from \eqref{eq:lip_eom}, then solving for $F_x$ and $F_y$ in \eqref{eq:planar_eom}. 

\section{MPC with QP}
\label{sec:mpc}

The LTI system shown in \eqref{eq:ss_lip} is used to formulate a QP problem to be utilized in an MPC. The equation of motion in \eqref{eq:ss_lip} is discretized into the following form:
\begin{equation}
\begin{gathered}
    \bm X_{k+1} = \bm F \, \bm X_k + \bm G \, \bm u_k \\
    \bm F = \bm I + \bm A\, \Delta t, \quad \bm G = \bm B\, \Delta t
\end{gathered}
\end{equation}
where $\Delta t$ is the MPC prediction time step. We derive the QP to track a desired reference $\bm X_{ref,k}$ with a prediction horizon $n_h$. Let $\bm Z = [\bm X_1^\top, \dots, \bm X_{n_h}^\top]^\top$ be the concatenated states in the prediction horizon and $\bm U = [\bm u_0^\top, \dots, \bm u_{n_h-1}^\top]^\top$ be the decision variable for the QP solver. Following the LTI system, the states within the prediction horizon can be derived as follows:
\begin{equation}
\begin{gathered}
    \bm Z = \bm H \, \bm U + \bm W \, \bm X_0 \\
    \bm W = \begin{bmatrix}
        \bm F^\top & (\bm F^2)^\top & \dots & (\bm F^{n_h})^\top
    \end{bmatrix}^\top \\
    \bm H = \begin{bmatrix}
        \bm G & \bm 0 & \bm 0 & \dots & \bm 0 \\
        \bm F \bm G & \bm G & \bm 0 & \dots & \bm 0 \\
        \vdots & \vdots & \vdots & \vdots & \vdots \\
        \bm F^{n_h-1} \bm G & \bm F^{n_h-2}\bm G & \bm F^{n_h-2} \bm G & \dots & \bm G
    \end{bmatrix}
\end{gathered}
\end{equation}
\begin{figure}
    \centering
    \includegraphics[width=\linewidth]{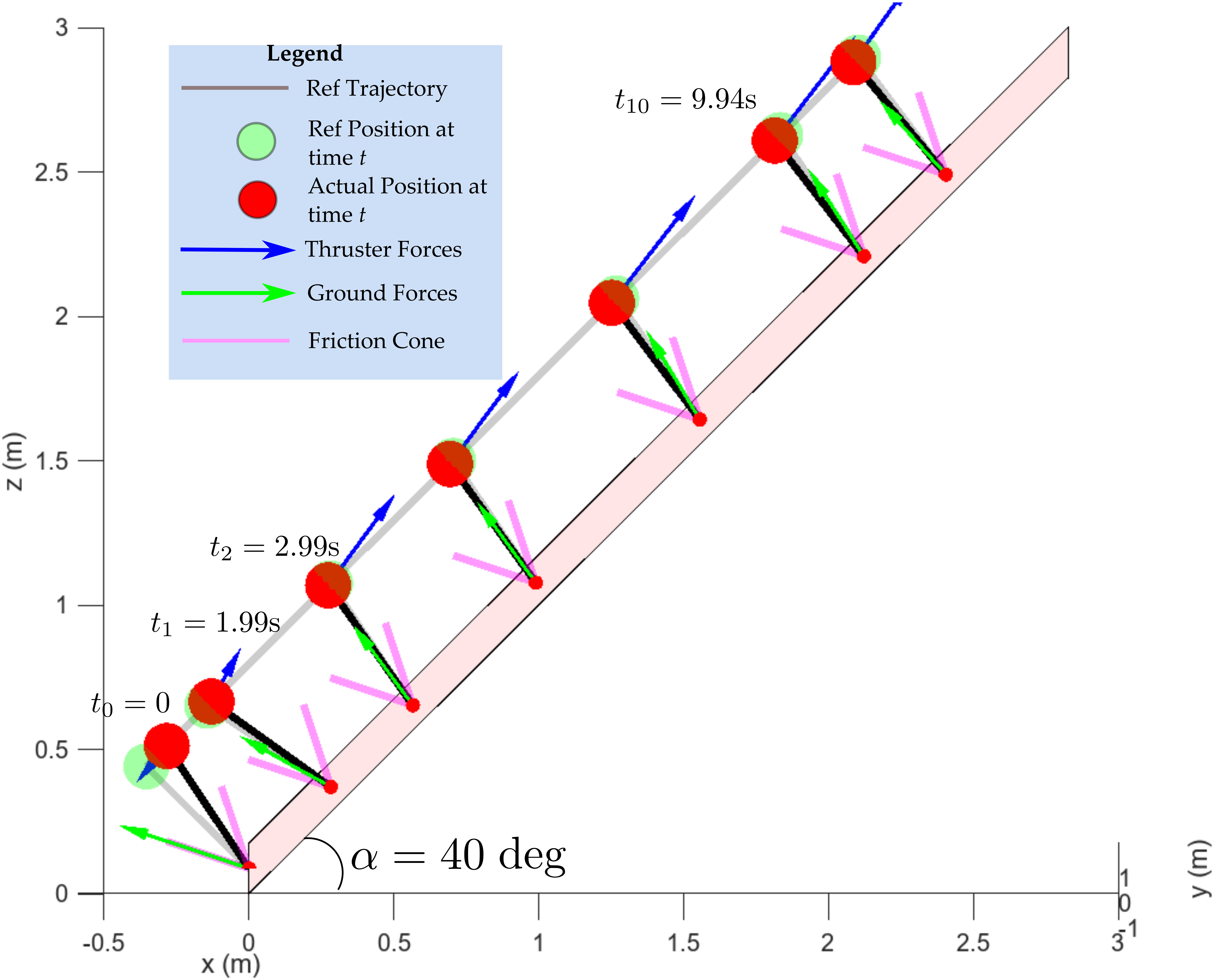}
    \caption{Figure depicting Snapshots from the Matlab simulation of the WAIR VLIP model going up a slope of 40 degrees}
    \label{fig:enter-label}
\end{figure}
The QP cost function to be solved is defined as the tracking error $J = \left\Vert \bm Z - \bm Z_{ref} \right\Vert_{\bm Q}^2$, which can be derived into the following generalized QP cost function:
\begin{equation}
\begin{aligned}
    J &= \bm U^\top \bm R \bm U + \bm b^\top \bm U + \bm c
\end{aligned}
\end{equation}
where $\bm R$, $\bm b$, and $\bm c$ are functions of $\bm Q$, $\bm H$, $\bm W$, and $\bm X_0$. Then, the optimization problem for the MPC can be derived as follows:
\begin{equation}
\begin{aligned}
    \min_{\bm U} \quad & J \\
    \mathrm{s.t.} \quad 
    & \bm U_{min} \leq \bm U \leq \bm U_{max} \\
    & \bm U \in \mathrm{FrictionCone}
\end{aligned}
\end{equation}
where $\bm U_{min}$ and $\bm U_{min}$ are bounds of $\bm U$, and the friction cone constraints are:
\begin{equation}
\begin{aligned}
    \lambda_{y,k} & > \lambda_{min_n}, \quad &
    \lambda_{x,k} & < \mu_s\,|\lambda_{y,k}|,
\end{aligned}
\end{equation}
where $\lambda_{min_n}$ is the minimum normal force desired. Then, the MPC will use $\bm u_0$ as the controller input in the simulation.

The primary advantage of this QP-based framework is its ability to adjust contact forces in both the front and hind limbs of the Husky, enabling placement of the CoP at arbitrary locations, including the midpoint of the support polygon formed by the contact points. Consequently, the thruster forces derived from the QP for the VLIP model can be seamlessly transferred to the full dynamics.

\section{Simulation Results and Discussion}

\begin{figure}
\centering
\includegraphics[width=1\linewidth]{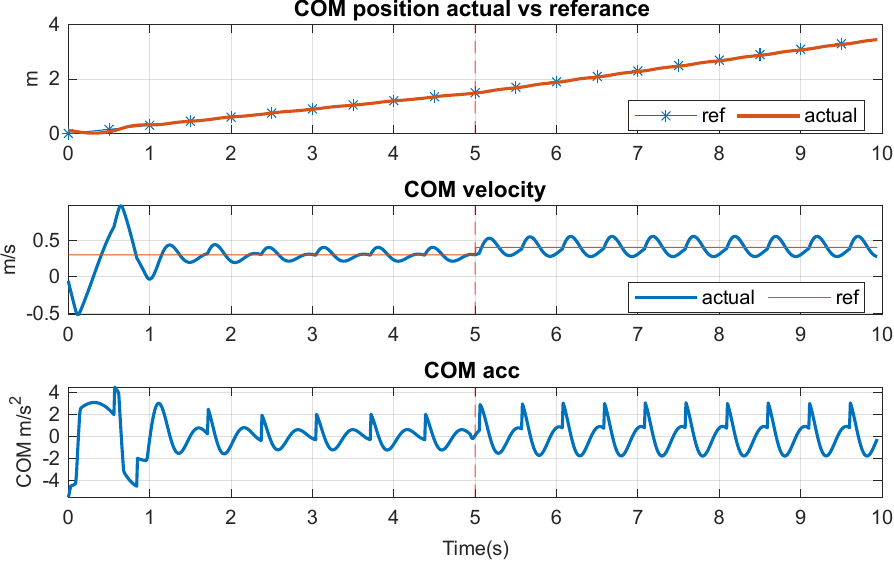}
    \caption{Error and tracking performance of body position and velocity}
    \label{fig:Body_states}
\end{figure}

\begin{figure}
    \centering   \includegraphics[width=\linewidth]{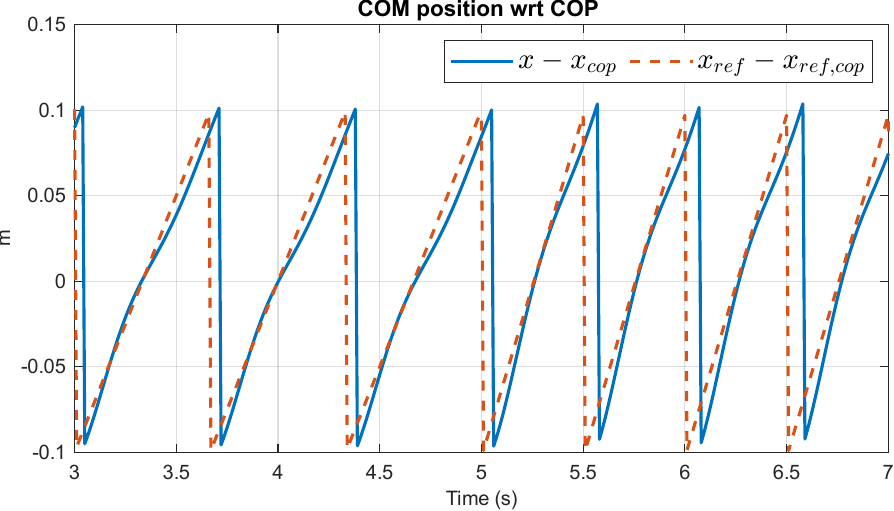}
    \caption{Position tracking in COP frame }
    \label{fig:Error_cop}
\end{figure}

\begin{figure}
    \centering  \includegraphics[width=1\linewidth]{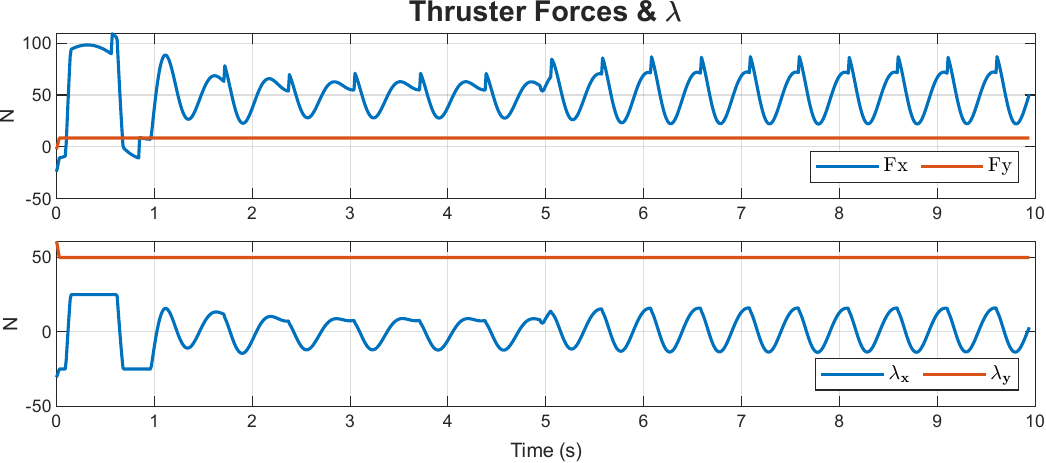}
    \caption{Calculated optimal ground reaction forces and thruster forces from \eqref{eq:eom}}
    \label{fig:forces_grouped}
\end{figure}

\begin{figure}
    \centering  \includegraphics[width=0.9\linewidth]{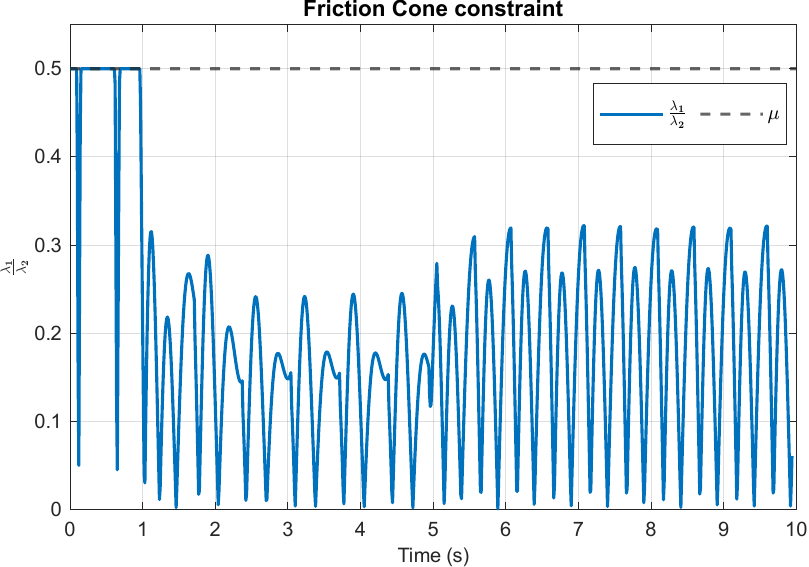}
    \caption{Friction cone constraint satisfaction of the ground reaction forces for a $\mu = 0.5$ }
    \label{fig:friction_constraint}
\end{figure}

\begin{figure}
    \centering   \includegraphics[width=0.9\linewidth]{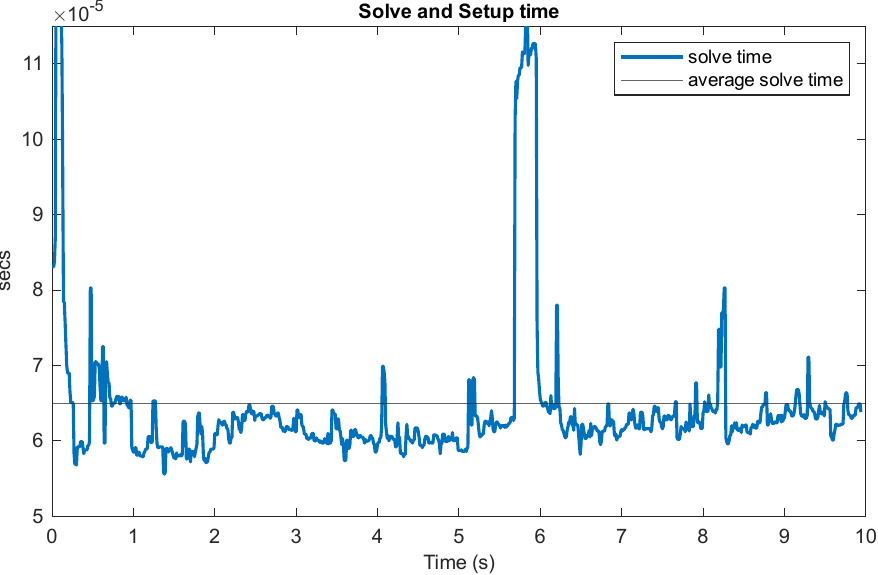}
    \caption{qpSWIFT solve times for a MPC horizon of 5 and step time of 0.01 seconds}
    \label{fig:solve-time}
\end{figure}

This simulation was performed in the MATLAB environment using a computer with an Intel core i7 processor and utilized the VLIP framework explained in Section \ref{sec:LIP_model}, supported by MATLAB animations, to model and analyze the system’s behavior. 

The simulation uses a 4th-order Runge-Kutta algorithm to march the states forward. For the purposes of this simulation, frontal plane was constrained and we only looked at the planar model in the sagittal plane.
A reference trajectory along the slope was generated and was rotated based on the slope of the incline. Fixed footholds, which translate to a desired fixed center of pressure point, are chosen based on a simple heuristic, which also defined how the center of pressure point was updated. For the sake of this simulation, this update rule was chosen so that the movement of the body is symmetrical with respect to the COP point.

The MPC, using the linearized of the modified VLIP dynamics, was able to consistently find feasible, and optimal ground reaction forces, as seen in Fig. ~\ref{fig:forces_grouped} and Fig.~\ref{fig:friction_constraint}, for a nominal ground friction coefficient of 0.5. The resultant thruster forces are also shown in Fig.~\ref{fig:forces_grouped}. As discussed in Section ~\ref{sec:LIP_model}, the ground reaction forces are not a function of the slope. It can be observed that the results are consistent with the intuition behind WAIR \cite{dial_wing-assisted_2003}. 

The required thruster forces are two dimensional and thruster forces normal to the plane of the slope allow to modulate the required normal ground force. Since acceleration normal to the plane is considered to be zero, the optimal normal ground reaction force from the optimizer, and consequently the thruster forces, form a constant value which is dependant on the $\lambda_{min_n}$. The optimizer tries to find the smallest possible ground normal force because of the inherent formulation of the quadratic cost function. 

The resultant trajectories are also able to track the reference position and velocity in the world frame in Fig.~\ref{fig:Body_states} which shows the position and velocity tracking in the world frame. Figure~\ref{fig:Error_cop} shows the same tracking performance parallel to the slope direction in the fixed COP frame. At $t = 5$ secs the reference trajectory steps up and the controller is able to track the desired velocity.
The QP solver \textit{qpSWIFT} is able to find solutions at a consistent rate with the solve times averaging out at 65 microseconds with a maximum number iterations at 10, which demonstrates capability of this controller formulation with paired with the tool to work on real-time machines. 
\section{Conclusions and Future Work}
The WAIR maneuver was observed on young gallinaceous birds that rely heavily on their legs and wings to perform near vertical wall climbs. This study, being inspired by this maneuver, employed a reduced order modified VLIP model along with an MPC to realize this on legged-aerial robots, specifically on a quadrupedal called Husky. The simulations indicate that it is possible to find the required optimal forces such that constraints are satisfied for this reduced-order model. These promising initial results provides insight on how these methods can be used, with the combination of the reduced order model and the MPC, to solve for thruster forces in varying terrains. 

The future pathways created by this research include finding ways to increase the collaboration between the legged and aerial modalities of the robot. This could involve modeling 3-dimensional movements by considering accelerations in all directions. Additionally, an effort to formulate a whole-body controller that modulates the ground reaction forces at the individual legs by controlling the joint accelerations would help obtain the required ground forces from the MPC at an arbitrarily fixed COP point.  

Eventually to try this on the actual hardware, strictly implementing the kinematic and dynamics constraints would be required. This would involve, finding and enforcing body pitch and roll angles to vector the desired thrust forces found by the controller. Considering the hybrid and cyclical nature of walking, which involves impulse forces and convergences to stable periodic motions, a hybrid model that could work with an appropriate model predictive controller, and is robust against noise and disturbances is also envisioned. To actually be able to direct the total thrust forces in the direction required a method of thrust regeneration using baffles and/or posture manipulation is being looked into.

\printbibliography

@misc{krishnamurthy_thruster-assisted_2024,
	title = {Thruster-{Assisted} {Incline} {Walking}},
	url = {http://arxiv.org/abs/2406.13118},
	doi = {10.48550/arXiv.2406.13118},
	urldate = {2024-08-26},
	publisher = {arXiv},
	author = {Krishnamurthy, Kaushik Venkatesh and Wang, Chenghao and Pitroda, Shreyansh and Salagame, Adarsh and Sihite, Eric and Nemovi, Reza and Ramezani, Alireza and Gharib, Morteza},
	month = jun,
	year = {2024},
	note = {arXiv:2406.13118 [cs, eess]},
}

@misc{salagame_quadrupedal_2023,
	title = {Quadrupedal {Locomotion} {Control} {On} {Inclined} {Surfaces} {Using} {Collocation} {Method}},
	url = {http://arxiv.org/abs/2312.08621},
	doi = {10.48550/arXiv.2312.08621},
	urldate = {2024-01-17},
	publisher = {arXiv},
	author = {Salagame, Adarsh and Gianello, Maria and Wang, Chenghao and Venkatesh, Kaushik and Pitroda, Shreyansh and Rajput, Rohit and Sihite, Eric and Leeser, Miriam and Ramezani, Alireza},
	month = dec,
	year = {2023},
	note = {arXiv:2312.08621 [cs, eess]},
}

@inproceedings{liang_rough-terrain_2021,
	title = {Rough-{Terrain} {Locomotion} and {Unilateral} {Contact} {Force} {Regulations} {With} a {Multi}-{Modal} {Legged} {Robot}},
	doi = {10.23919/ACC50511.2021.9483189},
	booktitle = {2021 {American} {Control} {Conference} ({ACC})},
	author = {Liang, Kaier and Sihite, Eric and Dangol, Pravin and Lessieur, Andrew and Ramezani, Alireza},
	month = may,
	year = {2021},
	note = {ISSN: 2378-5861},
	pages = {1762--1769},
}

@inproceedings{krishnamurthy_narrow-path_2024,
	title = {Narrow-{Path}, {Dynamic} {Walking} {Using} {Integrated} {Posture} {Manipulation} and {Thrust} {Vectoring}},
	url = {https://ieeexplore.ieee.org/document/10637015},
	doi = {10.1109/AIM55361.2024.10637015},
	urldate = {2024-08-26},
	booktitle = {2024 {IEEE} {International} {Conference} on {Advanced} {Intelligent} {Mechatronics} ({AIM})},
	author = {Krishnamurthy, Kaushik Venkatesh and Wang, Chenghao and Pitroda, Shreyansh and Salagame, Adarsh and Sihite, Eric and Nemovi, Reza and Ramezani, Alireza and Gharib, Morteza},
	month = jul,
	year = {2024},
	note = {ISSN: 2159-6255},
	pages = {898--903},
}

@inproceedings{sihite_optimization-free_2021,
	title = {Optimization-free {Ground} {Contact} {Force} {Constraint} {Satisfaction} in {Quadrupedal} {Locomotion}},
	doi = {10.1109/CDC45484.2021.9683155},
	booktitle = {2021 60th {IEEE} {Conference} on {Decision} and {Control} ({CDC})},
	author = {Sihite, Eric and Dangol, Pravin and Ramezani, Alireza},
	month = dec,
	year = {2021},
	note = {ISSN: 2576-2370},
	pages = {713--719},
}

@article{sihite_multi-modal_2023,
	title = {Multi-{Modal} {Mobility} {Morphobot} ({M4}) with appendage repurposing for locomotion plasticity enhancement},
	volume = {14},
	copyright = {2023 Springer Nature Limited},
	issn = {2041-1723},
	url = {https://www.nature.com/articles/s41467-023-39018-y},
	doi = {10.1038/s41467-023-39018-y},
	language = {en},
	number = {1},
	urldate = {2023-10-07},
	journal = {Nature Communications},
	author = {Sihite, Eric and Kalantari, Arash and Nemovi, Reza and Ramezani, Alireza and Gharib, Morteza},
	month = jun,
	year = {2023},
	note = {Number: 1
Publisher: Nature Publishing Group},
	pages = {3323},
}

@inproceedings{mandralis_minimum_2023,
	title = {Minimum {Time} {Trajectory} {Generation} for {Bounding} {Flight}: {Combining} {Posture} {Control} and {Thrust} {Vectoring}},
	shorttitle = {Minimum {Time} {Trajectory} {Generation} for {Bounding} {Flight}},
	url = {https://ieeexplore.ieee.org/document/10178360},
	doi = {10.23919/ECC57647.2023.10178360},
	urldate = {2024-08-26},
	booktitle = {2023 {European} {Control} {Conference} ({ECC})},
	author = {Mandralis, Ioannis and Sihite, Eric and Ramezani, Alireza and Gharib, Morteza},
	month = jun,
	year = {2023},
	pages = {1--7},
}

@inproceedings{ramezani_generative_2021,
	title = {Generative {Design} of {NU}’s {Husky} {Carbon}, {A} {Morpho}-{Functional}, {Legged} {Robot}},
	url = {https://ieeexplore.ieee.org/abstract/document/9561196},
	doi = {10.1109/ICRA48506.2021.9561196},
	urldate = {2023-11-22},
	booktitle = {2021 {IEEE} {International} {Conference} on {Robotics} and {Automation} ({ICRA})},
	author = {Ramezani, Alireza and Dangol, Pravin and Sihite, Eric and Lessieur, Andrew and Kelly, Peter},
	month = may,
	year = {2021},
	note = {ISSN: 2577-087X},
	pages = {4040--4046},
}

@inproceedings{sihite_demonstrating_2023,
	title = {Demonstrating {Autonomous} {3D} {Path} {Planning} on a {Novel} {Scalable} {UGV}-{UAV} {Morphing} {Robot}},
	url = {https://ieeexplore.ieee.org/document/10342189},
	doi = {10.1109/IROS55552.2023.10342189},
	urldate = {2024-08-26},
	booktitle = {2023 {IEEE}/{RSJ} {International} {Conference} on {Intelligent} {Robots} and {Systems} ({IROS})},
	author = {Sihite, Eric and Slezak, Filip and Mandralis, Ioannis and Salagame, Adarsh and Ramezani, Milad and Kalantari, Arash and Ramezani, Alireza and Gharib, Morteza},
	month = oct,
	year = {2023},
	note = {ISSN: 2153-0866},
	pages = {3064--3069},
}

@misc{salagame_letter_2022,
	title = {A {Letter} on {Progress} {Made} on {Husky} {Carbon}: {A} {Legged}-{Aerial}, {Multi}-modal {Platform}},
	shorttitle = {A {Letter} on {Progress} {Made} on {Husky} {Carbon}},
	url = {http://arxiv.org/abs/2207.12254},
	doi = {10.48550/arXiv.2207.12254},
	urldate = {2023-05-17},
	publisher = {arXiv},
	author = {Salagame, Adarsh and Manjikian, Shoghair and Wang, Chenghao and Krishnamurthy, Kaushik Venkatesh and Pitroda, Shreyansh and Gupta, Bibek and Jacob, Tobias and Mottis, Benjamin and Sihite, Eric and Ramezani, Milad and Ramezani, Alireza},
	month = jul,
	year = {2022},
	note = {arXiv:2207.12254 [cs, eess]},
}

@article{kim_bipedal_2021,
	title = {A bipedal walking robot that can fly, slackline, and skateboard},
	volume = {6},
	url = {https://www.science.org/doi/full/10.1126/scirobotics.abf8136},
	doi = {10.1126/scirobotics.abf8136},
	number = {59},
	urldate = {2023-05-17},
	journal = {Science Robotics},
	author = {Kim, Kyunam and Spieler, Patrick and Lupu, Elena-Sorina and Ramezani, Alireza and Chung, Soon-Jo},
	month = oct,
	year = {2021},
	note = {Publisher: American Association for the Advancement of Science},
	pages = {eabf8136},
}

@inproceedings{peterson_experimental_2011,
	title = {Experimental dynamics of wing assisted running for a bipedal ornithopter},
	doi = {10.1109/IROS.2011.6095041},
	booktitle = {2011 {IEEE}/{RSJ} {International} {Conference} on {Intelligent} {Robots} and {Systems}},
	author = {Peterson, Kevin and Fearing, Ronald S.},
	month = sep,
	year = {2011},
	note = {ISSN: 2153-0866},
	pages = {5080--5086},
}

@article{dial_wing-assisted_2003,
	title = {Wing-{Assisted} {Incline} {Running} and the {Evolution} of {Flight}},
	volume = {299},
	url = {https://www.science.org/doi/10.1126/science.1078237},
	doi = {10.1126/science.1078237},
	number = {5605},
	urldate = {2022-10-05},
	journal = {Science},
	author = {Dial, Kenneth P.},
	month = jan,
	year = {2003},
	note = {Publisher: American Association for the Advancement of Science},
	pages = {402--404},
}

@article{tobalske_aerodynamics_2007,
	title = {Aerodynamics of wing-assisted incline running in birds},
	volume = {210},
	issn = {0022-0949},
	url = {https://doi.org/10.1242/jeb.001701},
	doi = {10.1242/jeb.001701},
	number = {10},
	urldate = {2022-10-03},
	journal = {Journal of Experimental Biology},
	author = {Tobalske, Bret W. and Dial, Kenneth P.},
	month = may,
	year = {2007},
	pages = {1742--1751},
}

@article{focchi_high-slope_2017,
	title = {High-slope terrain locomotion for torque-controlled quadruped robots},
	volume = {41},
	issn = {1573-7527},
	url = {https://doi.org/10.1007/s10514-016-9573-1},
	doi = {10.1007/s10514-016-9573-1},
	language = {en},
	number = {1},
	urldate = {2020-05-30},
	journal = {Autonomous Robots},
	author = {Focchi, Michele and del Prete, Andrea and Havoutis, Ioannis and Featherstone, Roy and Caldwell, Darwin G. and Semini, Claudio},
	month = jan,
	year = {2017},
	pages = {259--272},
}

@inproceedings{ding_real-time_2019,
	title = {Real-time {Model} {Predictive} {Control} for {Versatile} {Dynamic} {Motions} in {Quadrupedal} {Robots}},
	doi = {10.1109/ICRA.2019.8793669},
	booktitle = {2019 {International} {Conference} on {Robotics} and {Automation} ({ICRA})},
	author = {Ding, Y. and Pandala, A. and Park, H.},
	month = may,
	year = {2019},
	pages = {8484--8490},
}

@inproceedings{katz_mini_2019,
	title = {Mini {Cheetah}: {A} {Platform} for {Pushing} the {Limits} of {Dynamic} {Quadruped} {Control}},
	shorttitle = {Mini {Cheetah}},
	doi = {10.1109/ICRA.2019.8793865},
	booktitle = {2019 {International} {Conference} on {Robotics} and {Automation} ({ICRA})},
	author = {Katz, B. and Carlo, J. D. and Kim, S.},
	month = may,
	year = {2019},
	pages = {6295--6301},
}

@inproceedings{hutter_hybrid_2012,
	title = {Hybrid {Operational} {Space} {Control} for {Compliant} {Legged} {Systems}},
	isbn = {978-0-262-51968-7},
	url = {http://www.roboticsproceedings.org/rss08/p17.pdf},
	doi = {10.15607/RSS.2012.VIII.017},
	urldate = {2024-07-15},
	booktitle = {Robotics: {Science} and {Systems} {VIII}},
	publisher = {Robotics: Science and Systems Foundation},
	author = {Hutter, Marco and Hoepflinger, Mark and Gehring, Christian and Bloesch, Michael and David Remy, C. and Siegwart, Roland},
	month = jul,
	year = {2012},
}

@article{komatsu_how_2015,
	title = {How to optimize the slope walking motion by the quadruped walking robot},
	volume = {29},
	issn = {0169-1864},
	url = {https://doi.org/10.1080/01691864.2015.1081103},
	doi = {10.1080/01691864.2015.1081103},
	number = {23},
	urldate = {2024-07-15},
	journal = {Advanced Robotics},
	author = {Komatsu, Hirone and Endo, Gen and Hodoshima, Ryuichi and Hirose, Shigeo and Fukushima, Edwardo F.},
	month = dec,
	year = {2015},
	note = {Publisher: Taylor \& Francis
\_eprint: https://doi.org/10.1080/01691864.2015.1081103},
	pages = {1497--1509},
}

@inproceedings{hirose_titan_1997,
	title = {{TITAN} {VII}: quadruped walking and manipulating robot on a steep slope},
	volume = {1},
	shorttitle = {{TITAN} {VII}},
	url = {https://ieeexplore.ieee.org/abstract/document/620085?casa_token=Ay1q3zQRosQAAAAA:U6kt0kdM9mYp3S6rVdrdk5i30kdaiO36KjemabYnZwAZcAOPMK8u-d8lKKdq_dfsW0mY4B1QVS8},
	doi = {10.1109/ROBOT.1997.620085},
	urldate = {2024-07-15},
	booktitle = {Proceedings of {International} {Conference} on {Robotics} and {Automation}},
	author = {Hirose, S. and Yoneda, K. and Tsukagoshi, H.},
	month = apr,
	year = {1997},
	pages = {494--500 vol.1},
}

@article{pandala_qpswift_2019,
	title = {{qpSWIFT}: {A} {Real}-{Time} {Sparse} {Quadratic} {Program} {Solver} for {Robotic} {Applications}},
	volume = {4},
	issn = {2377-3766},
	shorttitle = {{qpSWIFT}},
	url = {https://ieeexplore.ieee.org/document/8754693},
	doi = {10.1109/LRA.2019.2926664},
	number = {4},
	urldate = {2024-07-11},
	journal = {IEEE Robotics and Automation Letters},
	author = {Pandala, Abhishek Goud and Ding, Yanran and Park, Hae-Won},
	month = oct,
	year = {2019},
	note = {Conference Name: IEEE Robotics and Automation Letters},
	pages = {3355--3362},
}

@inproceedings{roennau_reactive_2014,
	title = {Reactive posture behaviors for stable legged locomotion over steep inclines and large obstacles},
	url = {https://ieeexplore.ieee.org/document/6943257},
	doi = {10.1109/IROS.2014.6943257},
	urldate = {2024-07-11},
	booktitle = {2014 {IEEE}/{RSJ} {International} {Conference} on {Intelligent} {Robots} and {Systems}},
	author = {Roennau, A. and Heppner, G. and Nowicki, M. and Zoellner, J.M. and Dillmann, R.},
	month = sep,
	year = {2014},
	note = {ISSN: 2153-0866},
	pages = {4888--4894},
}

@article{peterson_wing-assisted_2011,
	title = {A wing-assisted running robot and implications for avian flight evolution},
	volume = {6},
	issn = {1748-3190},
	url = {https://dx.doi.org/10.1088/1748-3182/6/4/046008},
	doi = {10.1088/1748-3182/6/4/046008},
	language = {en},
	number = {4},
	urldate = {2024-07-11},
	journal = {Bioinspiration \& Biomimetics},
	author = {Peterson, K. and Birkmeyer, P. and Dudley, R. and Fearing, R. S.},
	month = oct,
	year = {2011},
	pages = {046008},
}

@inproceedings{ma_quadrupedal_2020,
	title = {Quadrupedal {Robotic} {Walking} on {Sloped} {Terrains} via {Exact} {Decomposition} into {Coupled} {Bipedal} {Robots}},
	url = {https://ieeexplore.ieee.org/document/9341181},
	doi = {10.1109/IROS45743.2020.9341181},
	urldate = {2024-07-11},
	booktitle = {2020 {IEEE}/{RSJ} {International} {Conference} on {Intelligent} {Robots} and {Systems} ({IROS})},
	author = {Ma, Wen-Loong and Csomay-Shanklin, Noel and Ames, Aaron D.},
	month = oct,
	year = {2020},
	note = {ISSN: 2153-0866},
	pages = {4006--4011},
}

@article{xin_optimization-based_2024,
	title = {Optimization-based dynamic motion planning and control for quadruped robots},
	volume = {112},
	issn = {1573-269X},
	url = {https://doi.org/10.1007/s11071-024-09445-7},
	doi = {10.1007/s11071-024-09445-7},
	language = {en},
	number = {9},
	urldate = {2024-07-01},
	journal = {Nonlinear Dynamics},
	author = {Xin, Guiyang and Mistry, Michael},
	month = may,
	year = {2024},
	pages = {7043--7056},
}

@misc{salagame_quadrupedal_2023-1,
	title = {Quadrupedal {Locomotion} {Control} {On} {Inclined} {Surfaces} {Using} {Collocation} {Method}},
	url = {http://arxiv.org/abs/2312.08621},
	doi = {10.48550/arXiv.2312.08621},
	urldate = {2024-04-05},
	publisher = {arXiv},
	author = {Salagame, Adarsh and Gianello, Maria and Wang, Chenghao and Venkatesh, Kaushik and Pitroda, Shreyansh and Rajput, Rohit and Sihite, Eric and Leeser, Miriam and Ramezani, Alireza},
	month = dec,
	year = {2023},
	note = {arXiv:2312.08621 [cs, eess]},
}

@inproceedings{gehring_dynamic_2015,
	title = {Dynamic trotting on slopes for quadrupedal robots},
	url = {https://ieeexplore.ieee.org/abstract/document/7354099},
	doi = {10.1109/IROS.2015.7354099},
	urldate = {2024-03-23},
	booktitle = {2015 {IEEE}/{RSJ} {International} {Conference} on {Intelligent} {Robots} and {Systems} ({IROS})},
	author = {Gehring, Christian and Bellicoso, C. Dario and Coros, Stelian and Bloesch, Michael and Fankhauser, Péter and Hutter, Marco and Siegwart, Roland},
	month = sep,
	year = {2015},
	pages = {5129--5135},
}

@article{ding_representation-free_2021,
	title = {Representation-{Free} {Model} {Predictive} {Control} for {Dynamic} {Motions} in {Quadrupeds}},
	volume = {37},
	issn = {1552-3098, 1941-0468},
	url = {http://arxiv.org/abs/2012.10002},
	doi = {10.1109/TRO.2020.3046415},
	number = {4},
	urldate = {2023-12-18},
	journal = {IEEE Transactions on Robotics},
	author = {Ding, Yanran and Pandala, Abhishek and Li, Chuanzheng and Shin, Young-Ha and Park, Hae-Won},
	month = aug,
	year = {2021},
	note = {arXiv:2012.10002 [cs]},
	pages = {1154--1171},
}

@misc{sihite_dynamic_2023,
	title = {Dynamic modeling of wing-assisted inclined running with a morphing multi-modal robot},
	url = {http://arxiv.org/abs/2311.09963},
	doi = {10.48550/arXiv.2311.09963},
	urldate = {2023-12-09},
	publisher = {arXiv},
	author = {Sihite, Eric and Ramezani, Alireza and Gharib, Morteza},
	month = nov,
	year = {2023},
	note = {arXiv:2311.09963 [cs, eess]},
}

@inproceedings{griffin_walking_2017,
	title = {Walking stabilization using step timing and location adjustment on the humanoid robot, {Atlas}},
	url = {https://ieeexplore.ieee.org/abstract/document/8202223},
	doi = {10.1109/IROS.2017.8202223},
	urldate = {2023-11-27},
	booktitle = {2017 {IEEE}/{RSJ} {International} {Conference} on {Intelligent} {Robots} and {Systems} ({IROS})},
	author = {Griffin, Robert J. and Wiedebach, Georg and Bertrand, Sylvain and Leonessa, Alexander and Pratt, Jerry},
	month = sep,
	year = {2017},
	note = {ISSN: 2153-0866},
	pages = {667--673},
}

@inproceedings{lee_enhanced_2023,
	title = {Enhanced {Balance} for {Legged} {Robots} {Using} {Reaction} {Wheels}},
	url = {https://ieeexplore.ieee.org/document/10160833},
	doi = {10.1109/ICRA48891.2023.10160833},
	urldate = {2023-11-20},
	booktitle = {2023 {IEEE} {International} {Conference} on {Robotics} and {Automation} ({ICRA})},
	author = {Lee, Chi-Yen and Yang, Shuo and Bokser, Benjamin and Manchester, Zachary},
	month = may,
	year = {2023},
	pages = {9980--9987},
}

\end{document}